\documentclass{article} % For LaTeX2e

\usepackage{iclr2016_conference,times}
\usepackage{hyperref}
\usepackage{url}

\usepackage{graphicx}
\graphicspath{ {fig/} }
\usepackage{subfigure} 

\usepackage{natbib}
\usepackage{enumitem}

\usepackage{algorithm}
\usepackage{algorithmic}

\usepackage{amsmath}
\usepackage{amssymb}

\usepackage[toc,page]{appendix}
\numberwithin{equation}{section}

\usepackage{tikz}
\usetikzlibrary{er,positioning,shapes,snakes,shadows,arrows, fit, topaths,matrix,arrows} %arrows.meta}
\makeatletter
\pgfkeys{/pgf/.cd,
  parallelepiped offset x/.initial=2mm,
  parallelepiped offset y/.initial=2mm
}
\pgfdeclareshape{parallelepiped}
{
  \inheritsavedanchors[from=rectangle] % this is nearly a rectangle
  \inheritanchorborder[from=rectangle]
  \inheritanchor[from=rectangle]{north}
  \inheritanchor[from=rectangle]{north west}
  \inheritanchor[from=rectangle]{north east}
  \inheritanchor[from=rectangle]{center}
  \inheritanchor[from=rectangle]{west}
  \inheritanchor[from=rectangle]{east}
  \inheritanchor[from=rectangle]{mid}
  \inheritanchor[from=rectangle]{mid west}
  \inheritanchor[from=rectangle]{mid east}
  \inheritanchor[from=rectangle]{base}
  \inheritanchor[from=rectangle]{base west}
  \inheritanchor[from=rectangle]{base east}
  \inheritanchor[from=rectangle]{south}
  \inheritanchor[from=rectangle]{south west}
  \inheritanchor[from=rectangle]{south east}
  \backgroundpath{
    \southwest \pgf@xa=\pgf@x \pgf@ya=\pgf@y
    \northeast \pgf@xb=\pgf@x \pgf@yb=\pgf@y
    \pgfmathsetlength\pgfutil@tempdima{\pgfkeysvalueof{/pgf/parallelepiped offset x}}
    \pgfmathsetlength\pgfutil@tempdimb{\pgfkeysvalueof{/pgf/parallelepiped offset y}}
    \def\ppd@offset{\pgfpoint{\pgfutil@tempdima}{\pgfutil@tempdimb}}
    \pgfpathmoveto{\pgfqpoint{\pgf@xa}{\pgf@ya}}
    \pgfpathlineto{\pgfqpoint{\pgf@xb}{\pgf@ya}}
    \pgfpathlineto{\pgfqpoint{\pgf@xb}{\pgf@yb}}
    \pgfpathlineto{\pgfqpoint{\pgf@xa}{\pgf@yb}}
    \pgfpathclose
    \pgfpathmoveto{\pgfqpoint{\pgf@xb}{\pgf@ya}}
    \pgfpathlineto{\pgfpointadd{\pgfpoint{\pgf@xb}{\pgf@ya}}{\ppd@offset}}
    \pgfpathlineto{\pgfpointadd{\pgfpoint{\pgf@xb}{\pgf@yb}}{\ppd@offset}}
    \pgfpathlineto{\pgfpointadd{\pgfpoint{\pgf@xa}{\pgf@yb}}{\ppd@offset}}
    \pgfpathlineto{\pgfqpoint{\pgf@xa}{\pgf@yb}}
    \pgfpathmoveto{\pgfqpoint{\pgf@xb}{\pgf@yb}}
    \pgfpathlineto{\pgfpointadd{\pgfpoint{\pgf@xb}{\pgf@yb}}{\ppd@offset}}
  }
}
\makeatother

\title{Symmetries and  control in generative neural nets.}
\author{Galin Georgiev \\ %\thanks{www.gammadynamics.com} \\
GammaDynamics, LLC\\
\texttt{galin.georgiev@gammadynamics.com } 
}

\iclrfinalcopy % Uncomment for camera-ready version

\begin{document}

\maketitle

\begin{abstract}
We study generative nets which can control and modify observations,  after being trained on real-life datasets. In order to zoom-in on an object, some spatial, color and other attributes are learned by classifiers in specialized \emph{attention} nets. In field-theoretical terms, these learned \emph{symmetry statistics} form the \emph{gauge group} of the data set. Plugging them in the generative layers of auto-classifiers-encoders (ACE) appears to be the most direct way to simultaneously: i) generate new observations with arbitrary attributes, from a given class; ii) describe the low-dimensional  manifold encoding the ``essence'' of the data, after superfluous attributes are factored out; and iii) organically control, i.e., move or modify objects within given observations. We demonstrate the sharp improvement of the generative qualities of shallow ACE, with added spatial and color symmetry statistics, on the distorted MNIST and CIFAR10 datasets. 
\end{abstract}

\section{Introduction}

\subsection{Generativity and control.}
\label{Generativity and}
Generating  plausible but unseen previously observations appears, at least chronologically,  to have been one of the hardest challenges for artificial neural nets. A generative net can ``dream-up''  new observations $\{\hat{\mathbf{x}}_{\nu}\}$, each a vector in a high-dimensional space $\mathbb{R}^{N}$, by sampling from a  white noise probability density $p(\mathbf{z})$. This \emph{model} density resides on a preferably low-dimensional space of \emph{latent} variables $\mathbf{z}$ = $\{\mathbf{z}^{(\kappa)}\}_{\kappa=1}^{N_{lat}}$. In order to create plausible new observations, the latent manifold  has to \emph{encode} the complexity of the set of $P$ training observations $\{\mathbf{x}_{\mu}\}_{\mu=1}^P$ $\subset \mathbb{R}^N$. 

Generativity has a lot more to it than  ``dreaming-up'' new random observations. It is at the heart of the control skills of a neural net. Visual biological nets, for example, capture existential motor information like location/shape and other attributes of an object and can act on it by moving or modifying it deterministically. Asking for this data compression to be as compact and low-dimensional as possible is therefore not only a general minimalist requirement.  Learning and mastering control is a gradual process, which naturally starts by seeking and exploring only a few degrees of freedom. 

Moreover, the ability to modify an object implies an ability to first and foremost reconstruct it, with various degrees of precision. Not unlike human creativity, a fully generative net has to balance out and minimize terms with non-compatible objectives: a) a \emph{generative} error term, which is responsible for converting random noise into plausible data, on the one hand, and b) a \emph{reconstruction} error term which is responsible for meticulous reconstruction of existing objects, on the other.

%%%%%%%%%%%%%%%%%%%%%%%%
\subsection{Learning from real-life data.}

\begin{figure}[!ht]
%\vskip 0.2in
\includegraphics[width=\columnwidth]{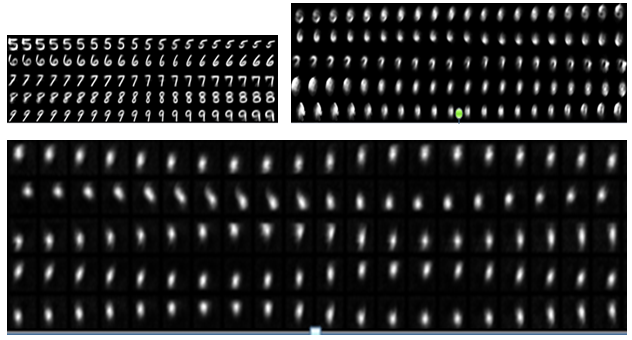}
\caption{One-dimensional latent manifold for some of the MNIST  classes, each row corresponding to a separate class. \textbf{Top Left.} Original MNIST, on 28x28 canvas. \textbf{Top Right.} RTS MNIST, on 42x42 canvas. \textbf{Bottom.} TC MNIST, on 60x60 canvas, Appendix \ref{Distorted MNIST}.
The net is a generative ACE in creative regime \cite{Georgiev15-0}. The latent layer is one-dimensional per class, traversed by an equally spaced  deterministic grid $\{\sigma_s \}_{s= 1}^{20}$, $-4 \leq \sigma_s \leq 4$. Implementation details in Appendix \ref{Implementation}.}
\label{Fig1.1}
%\vskip -0.2in
\end{figure}

From the recent crop of generative nets, section \ref{Generative nets}, only one appears to offer this desirable reconstruction via a low-dimensional latent manifold: the \emph{variational auto-encoders} (VAE) \cite{Kingma14-1}, \cite{Rezende14}. Their subset called \emph{Gibbs machines}, has also far-reaching roots into information geometry and thermodynamics, which come in very handy. They perform well on idealized visual data sets like MNIST \cite{LeCun98}. Unfortunately, like the other generative nets, they do not cope well with more realistic images, when objects are spatially varied or, if there is heavy clutter in the background. These traits are simulated in  the rotated-translated-scaled (RTS) MNIST and translated-cluttered (TC) MNIST, Appendix \ref{Distorted MNIST}. We highlight the shortcomings of basic generative nets on Figure \ref{Fig1.1}, for  the simplest case of one-dimensional latent manifold per class. While simulating wonderfully on the original MNIST (top-left), even with  $N_{lat} = 1$,  the net fails miserably to learn the distorted data: The  randomly ``dreamed-up'' samples $\{\hat{\mathbf{x}}_{\nu}\}$ are blurred and not plausible (top-right and bottom). Low latent dimensionality is not the culprit:  latent manifolds with dimensions $N_{lat} \geq 100$ do not yield much better results.

For  the real-life CIFAR10 dataset, \cite{Krizhevsky09}, the latent two-dimensional\footnote{The color scheme ``appropriates'' at least one latent dimension, hence the need for more dimensions.} manifold of the class of horses, produced by the same architecture, is on the left of Figure \ref{Fig1.3}. The training dataset has horses of different colors, facing both left and right, so the latent manifold tends to produce two-headed vague shapes of different colors.

%%%%%%%%%%%%%%%%%%%%%%%%
\subsection{``A horse, a horse! My kingdom for a horse!'' \protect\footnote{\cite{Shakespeare1592}}}
\label{A horse}
In order to get the horses back, we invoke  the Gibbs thermodynamic framework. It allows adding non-energy attributes to the sampling distribution and  modifying them, randomly or deterministically. These \emph{symmetry statistics}, like location, size, angle, color etc, are factored-out at the start and factored back-in at the end. The auto-classifier-encoder (ACE) net with symmetry statistics was suggested in \cite{Georgiev15-0} and detailed in section \ref{ACE with} here. The latent manifolds it produces, for the above three MNIST datasets, are on Figure \ref{Fig1.2}: 
\begin{figure}[!ht]
%\vskip 0.2in
\includegraphics[width=\columnwidth]{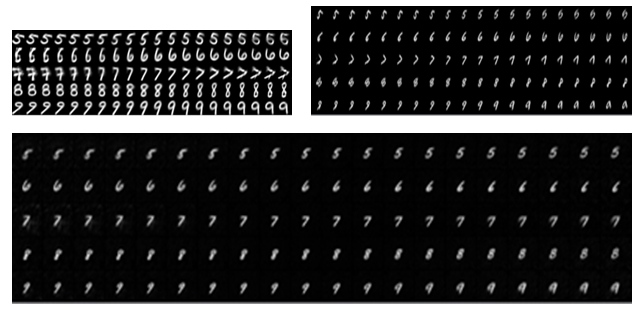}
\caption{The analog of Figure \ref{Fig1.1}, but produced by ACE with spatial symmetry statistics. For the original MNIST (top left), the size variation disappeared from the digit 5 class and the digit 7 class acquired a dash. In other words,  one sees more genuine ``core style'' variation, even with one latent dimension only. Implementation details in Appendix \ref{Implementation}.}
\label{Fig1.2}
%\vskip -0.2in
\end{figure} 
With distortions and clutter factored out, the quotient one-dimensional latent manifold is clear and legible. The  factorization is via transformations from the affine group $\mathbf{Aff}(2,\mathbb{R})$, which plays the role of the \emph{gauge group} in field theory. The \emph{spatial} symmetry statistics are the transformations parameters, computed via another optimizer net. The CIFAR10 horse class manifold, generated by ACE with spatial symmetry statistics, is on the right of Figure \ref{Fig1.3}. We have horse-like creatures, which morph into giraffes as one moves up the grid!
\begin{figure}[!h]
\begin{minipage}[]{0.5\columnwidth}
\includegraphics[width=\columnwidth]{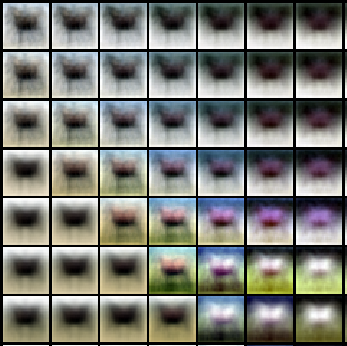}
\end{minipage} \hfill
\begin{minipage}[]{0.5\columnwidth}
\includegraphics[width=\columnwidth]{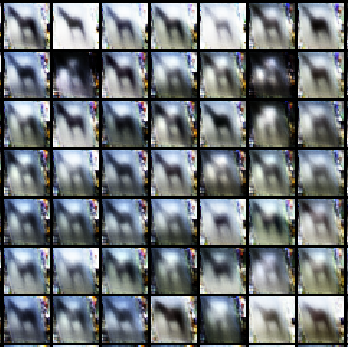}
\end{minipage}
\caption{\textbf{Left.} Latent manifold for the horse class in CIFAR10, using a shallow ACE, with two latent dimensions per class,  without symmetry statistics. These simulated images are from a 7x7 central segment of an equally spaced deterministic 30x30 grid $\{\sigma_s,\tau_s \}_{s= 1}^{30}$, $-6 \leq \sigma_s, \tau_s \leq 6$. \textbf{Right.} Same, but generated by shallow ACE \textbf{with spatial symmetry statistics} (implementation details in Appendix \ref{Implementation}). To  appreciate them, compare  to other generative nets:  Figure 2 (c) in \cite{Goodfellow14}, or Figure 3 (d) in \cite{Sohl15}. The  improvement in \cite{Denton15} is due to so-called  Laplacian pyramids, and can be overlayed on  any core generative model. }
\label{Fig1.3}
%\vskip -0.2in 
\end{figure}

\begin{figure}[!h]
\begin{minipage}[]{0.5\columnwidth}
\includegraphics[width=\columnwidth]{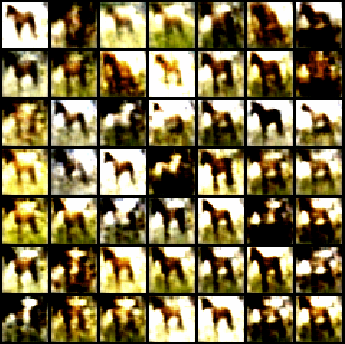}
\end{minipage} \hfill
\begin{minipage}[]{0.5\columnwidth}
\includegraphics[width=\columnwidth]{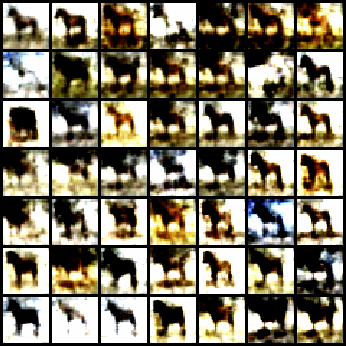}
\end{minipage}
\caption{Same as Figure \ref{Fig1.3}, but with the two fully-connected encoder hidden layers replaced by convolutional ones. Corresponding deconvolution layers, \cite{Zeiler14}, are added to  decoder.  \textbf{Left.} Shallow ACE with spatial symmetry statistics only. \textbf{Right.} Shallow ACE \textbf{with both spatial and color symmetry statistics}: as a result, the green background is  subdued. Implementation details in Appendix \ref{Implementation}.} 
\label{Fig1.4}
%\vskip -0.2in 
\end{figure}

The first successful application of Lie algebra symmetries to neural nets was in \cite{Simard00}. The recent crop of \emph{spatial attention} nets  \cite{Jadeberg15}, \cite{Gregor15}, \cite{Sermanet14}, \cite{Ba14} optimize spatial symmetry statistics, corresponding to a given object inside an observation. An efficient calculation of symmetry statistics, for multiple objects, requires a classifier. Hence, generation and reconstruction  on real-life datasets lead to  an auto-encoder/classifier combo like ACE. Supplementing auto-encoders with affine transforms was first proposed in \cite{Hinton11}, where spatial symmetry statistics were referred to as ``capsules''. As suggested there, hundreds and thousands of capsules can in principle be attached to feature maps. Current attention nets produce one set of symmetry statistics per object (inside an observation). Incorporating  convolutional feature maps in the encoder, and sampling from symmetry statistics at various depths, is yet to be engineered well for deep generative nets, see open problems \ref{Deconvolution}, \ref{Symm stats per feature}, section \ref{Open problems}. Results from a shallow convolutional ACE are on Figure \ref{Fig1.4}.   

Spatial symmetry statistics are vital in biological nets and can in principle be traced down experimentally. Feedback loops for attention data, vaguely reminiscent of Figure \ref{Fig3.1}, have  been identified between higher- and lower-level visual areas of the brain, \cite{Sherman05}%\footnote{The repeated feedback loop, relaying motor-related information between the cortex and the thalamus, and then to motor centers, is visualized on  Figures 5,6 there.}%
, \cite{Buffalo10}. 

For colored images, one also needs  the \emph{color symmetry statistics}, forming a semigroup of non-negative 3x3 matrices in the stochastic group\footnote{The subgroup of matrices $\in \mathbf{GL}(3, \mathbb{R})$, with entries in each row adding up to one, \cite{Poole95}.}  $\mathbf{S}(3,\mathbb{R})$. As shown on the right of Figure \ref{Fig1.4}, they help subdue the background color, and perhaps,  more. In particle physics parlance, three-dimensional color images are described by  \emph{chromodynamics} with a minimum gauge group  $\mathbf{Aff}(3,\mathbb{R})$ $\times ~ \mathbf{S}(3,\mathbb{R})$. 

The rest of the paper is organized as follows: section \ref{Generative nets}   briefly overviews recent generative nets and details VAE-s objective function; section \ref{Theoretical framework}  outlines  the theoretical framework of generative nets with control, highlighting the connections with information geometry and thermodynamics; section \ref{ACE with} presents the enhanced ACE architecture; the Appendices offer implementation and dataset details.

%%%%%%%%%%%%%%%%%%%%%%%%%%%%%%%%%%%%%%%%%%%%%%%%%%%%%%%%
\section{Generative nets and the latent manifold.}
\label{Generative nets}
Latent manifold learning was pioneered for modern nets in \cite{Rifai12}. When a latent sample $\mathbf {z}_{\nu}$ is chosen from a  model density $p(\mathbf{z})$, a generative net \emph{decodes} it into a simulated  observation $\hat{\mathbf{x}}_{\nu}$, from a corresponding model density $q(\hat{\mathbf{x}})$. There are two scenarios:

a) {\it the net has reconstruction capabilities}, hence $q(\mathbf{x})$ can in theory be evaluated on the training and testing observations $\{\mathbf{x}_{\mu}\}$. The objective is to minimize the so-called \emph{cross-entropy} or  \emph{negative log-likelihood}, i.e., the expectation $\mathbf{E}(- \log q(\mathbf{x}))_{r(\mathbf{x})}$, where $\mathbf{E}()_{r()}$ is an expectation with respect to the empirical density $r()$. Recently proposed reconstructive generative nets are: i) the generalized denoising auto-encoders (DAE) \cite{Bengio13}, ii) the generative stochastic networks (GSN) \cite{Bengio14}, iii) the variational auto-encoders introduced above, iv) the non-linear independent component estimation (NICE) \cite{Dinh14}, and v) \cite{Sohl15}. Except for NICE, the log-likelihood can not be exactly evaluated in practice, and is hence approximated. The first two models proxy  $q(\mathbf{x})$ with a certain conditional density $q(\mathbf{x} | \tilde{\mathbf{x}})$ and a Markov chain for the \emph{corrupted data} $\tilde{\mathbf{x}}$.  The variational auto-encoders proxy  the negative log-likelihood by a variational upper bound $\mathcal{U}(-\log q(\mathbf{x}))$. Method v) conjures up a forward diffusion process from $q(\mathbf{x})$ to $p(\mathbf{z})$ and uses the backward diffusion process to ``dream-up'' new observations $\{\hat{\mathbf{x}}_{\nu} \}$.

b) {\it the net has no reconstruction capabilities}, hence one has to resort to an interpolation  $q(\hat{\mathbf{x}})$  $\rightarrow$ $\hat{q}(\mathbf{x})$, in order to evaluate $q()$  on the training and testing observations $\{\mathbf{x}_{\mu}\}$. The objective is to minimize directly or indirectly the negative log-likelihood  $\mathbf{E}(- \log \hat{q}(\mathbf{x}))_{r(\mathbf{x})}$. Recent such model is the generative adversarial network (GAN) \cite{Goodfellow14}.   It minimizes indirectly the above negative log-likelihood by combining a generative and a discriminative net, the latter  tasked with distinguishing between the ``dreamed-up'' observations $\{\hat{\mathbf{x}}_{\nu} \}$ and training observations $\{\mathbf{x}_{\mu} \}$.

Of these models, only the variational auto-encoders and the generative adversarial networks are designed to handle a low-dimensional latent manifold. As argued in sub-section \ref{Generativity and}, reconstruction, i.e. scenario a), is an indispensable part of the  control skill set, hence we are left with the variational auto-encoder approach. As all generative nets, variational auto-encoders work in two regimes:
\begin{itemize}
\item \emph{creative} regime, with no data clamped onto the net and sampling from $p(\mathbf{z})$, and
\item \emph{non-creative} regime, with the training or testing observations $\{\mathbf{x}_{\mu}\}$   fed to the input layer of the net. Variational auto-encoders sample in this regime from a different closed-form conditional \emph{posterior} model density $p(\mathbf{z} | \mathbf{x}_{\mu})$.
\end{itemize}
In order to do reconstruction, variational auto-encoders also introduce a conditional model \emph{reconstruction} density $p^{rec}(\mathbf{x}_{\mu} | \mathbf{z})$. In non-creative regime, the reconstruction error at the output layer of the net is the expectation $\mathbf{E}(- \log p^{rec}(\mathbf{x}_{\mu}|\mathbf{z}))_{p(\mathbf{z}|\mathbf{x}_{\mu})}$. In the creative regime, we have a joint model density $p(\mathbf{x}_{\mu},\mathbf{z})$ $:=p^{rec}(\mathbf{x}_{\mu}|\mathbf{z})p(\mathbf{z})$. The data model density  $q(\mathbf{x}_{\mu})$ is the implied marginal: 
\begin{align}
q(\mathbf{x}_{\mu}) =\int p(\mathbf{x}_{\mu},\mathbf{z})d\mathbf{z} = \frac{ p(\mathbf{x}_{\mu},\mathbf{z})}{ q(\mathbf{z}|\mathbf{x}_{\mu}) },
\label{2.1}
\end{align}
for some implied posterior conditional density $q(\mathbf{z}|\mathbf{x}_{\mu})$ which  is generally intractable, $q(\mathbf{z}|\mathbf{x}_{\mu})$ $\neq p(\mathbf{z}|\mathbf{x}_{\mu})$. The full decomposition of our minimization target - the negative log-likelihood $-\log q(\mathbf{x}_{\mu})$ - is easily derived via the Bayes rules, \cite{Georgiev15-0}, section 3:
\begin{align}
-\log q(\mathbf{x}_{\mu}) =  \underbrace{\mathbf{E}(- \log p^{rec}(\mathbf{x}_{\mu}| \mathbf{z}))_{ p(\mathbf{z}|\mathbf{x}_{\mu})} }_{reconstruction~error} 
+ \underbrace{\mathcal{D}( p(\mathbf{z| x}_{\mu}) || p(\mathbf{z}))}_{generative~error} - \underbrace{\mathcal{D}( p(\mathbf{z| x}_{\mu}) || q(\mathbf{z| x}_{\mu}) )}_{variational~error},
\label{2.2}
\end{align}
where $\mathcal{D}( || )$ is the \emph{Kullback-Leibler divergence}. The \emph{reconstruction error} measures the negative likelihood of getting $\mathbf{x}_{\mu}$ back, after the transformations and  randomness inside the net. The \emph{generative error} is the  divergence between the generative densities in the non-creative and creative regimes. The \emph{variational error} is an approximation error: it is the price variational auto-encoders pay for having a tractable generative density $p(\mathbf{z} | \mathbf{x}_{\mu})$ in the non-creative regime. It is hard to compute, although some strides have been made, \cite{Rezende15}. For  the Gibbs machines discussed below, it was conjectured that  this error can be made arbitrary small, \cite{Georgiev15-0}. %, section 5, open problem 7.

%%%%%%%%%%%%%%%%%%%%%%%%%%%%%%%%%%%%%%%%%%%%%%%%%%%%%%%%%%%%%%%
\section{The theory. Connections with information geometry and thermodynamics.}
\label{Theoretical framework}
A theoretical framework for universal nets was recently outlined in \cite{Georgiev15-0}. Some of the constructs there, like the ACE architecture, appeared optional and driven solely by requirements for universality. We summarize and generalize the framework in the current context and argue that the ACE architecture, or its variations, are indispensable for generative  reconstructive nets. 
\begin{enumerate}

\item {\it Information geometry and Gibbs machines}: the minimization of the generative error in (\ref{2.2}) leads to sampling from Gibbs a.k.a. exponential class of densities. It follows from the probabilistic or variational Pythagorean theorem, \cite{Chentsov68}, which underlies modern estimation theory, and is pervasive in  information geometry, \cite{Amari00}. In the case of Laplacian \footnote{ The Laplacian density is not in the exponential class, but is a sum of two exponential densities which are in the exponential class in their respective domains.} generative densities, and conditionally independent latent variables $\mathbf{z}$ $=\{z^{(\kappa)}\}_{\kappa=1}^{N_{lat}}$, one has:
\begin{align}
p(\mathbf{z|x}_{\mu}) \sim   e^{-\sum_{\kappa=1}^{N_{lat}} p^{(\kappa)}_{\mu} |z^{(\kappa)} - m^{(\kappa)}_{\mu}| },
\label{2.3}
\end{align}
where the means $\{m_{\mu}^{(\kappa)} \}$ are symmetry statistics, the absolute value terms are \emph{sufficient statistics} and the inverse scale \emph{momenta} $\{p_{\mu}^{(\kappa)} \}$ are Lagrange multipliers, computed so as to satisfy given expectations of the sufficient statistics. The Gibbs density class leads to:

\item {\it Thermodynamics and  more symmetry statistics}: The Gibbs class is also central in thermodynamics because it is maximum-entropy class and allows to  add fluctuating attributes, other than energy. These additions are not cosmetic and fundamentally alter the dynamics of the canonical distribution, \cite{Landau80}, section 35. They can be any attributes: i) spatial attributes, as in the example below; ii) color attributes, as introduced in subsection \ref{A horse}, and others. For multiple objects, one needs specialized nets and a classifier to optimize them. This leads to:

\item {\it Auto-classifiers-encoder (ACE) architecture, section \ref{ACE with}}: Since classification labels are already needed above,  the latent manifold is better  learned: i) via supervised  reconstruction, and ii) with symmetry statistics used  by decoder. This leads to:

\item{\it Control}: With symmetry statistics in the generative layer, the net can organically move or modify the respective attributes of the objects, either deterministically or randomly. The ACE architecture ensures that the modifications stay within a given class.

\end{enumerate}
\textit{Example}: An important special case in visual recognition are  the spatial symmetry statistics, which describe the location, size, stance etc of an object. For a simple gray two-dimensional image $\mathbf{x}_{\mu}$ on $N$ pixels e.g., two of its spatial symmetry statistics are the coordinates $(h_{\mu}, v_{\mu})$ of its \emph{center of mass}, where the ''mass'' of a pixel is its intensity. Assuming independence, one can embed a translational invariance in the net, multiplying (\ref{2.3}) by the spatial symmetry statistics (SSS)  conditional density:
\begin{align}
p_{SSS}(\mathbf{z|x}_{\mu})  \sim  e^{-p^{(h)}_{\mu} |z^{(h)} - h_{\mu}| -p^{(v)}_{\mu} |z^{(v)} - v_{\mu}|},
\label{2.5}
\end{align}
where $z^{(h)}, z^{(v)}$ are two new zero-mean latent random variables, responsible respectively for horizontal and vertical translation. If $(\mathbf{h}, \mathbf{v})$ are the vectors of horizontal and vertical pixel coordinates, the image is centered at the input layer via the transform $(\mathbf{h}, \mathbf{v})$ $\rightarrow$ $(\mathbf{h} - h_{\mu}, \mathbf{v}-v_{\mu})$. This transformation is inverted, before reconstruction error is computed.

When rescaled and normalized, (\ref{2.5}) is the quantum mechanical probability density of a free particle, in imaginary space/time and  Planck constant  $\hslash=1$. Furthermore, for every observation $\mathbf{x}_{\mu}$, there could be multiple or infinitely many latents $\{\mathbf{z}_{\mu}^{(\kappa)} \}_{\kappa=1}^L$, $L \leq \infty$, and  $\mathbf{x}_{\mu}$ is merely a draw from a probability density $p^{rec}(\mathbf{x}_{\mu} | \mathbf{z})$. In a quantum statistics interpretation, latents are  microscopic quantum variables, while observables like pixels, are macroscopic aggregates. Observations represent \emph{partial equilibria} of independent small parts of the expanded (by a factor of $L$)  data set.

%%%%%%%%%%%%%%%%%%%%%%%%%%%%%%%%%%%%%%%%%%%%%%%%%%%%%%%%%%%
\section{ACE with symmetry statistics.}
\label{ACE with}

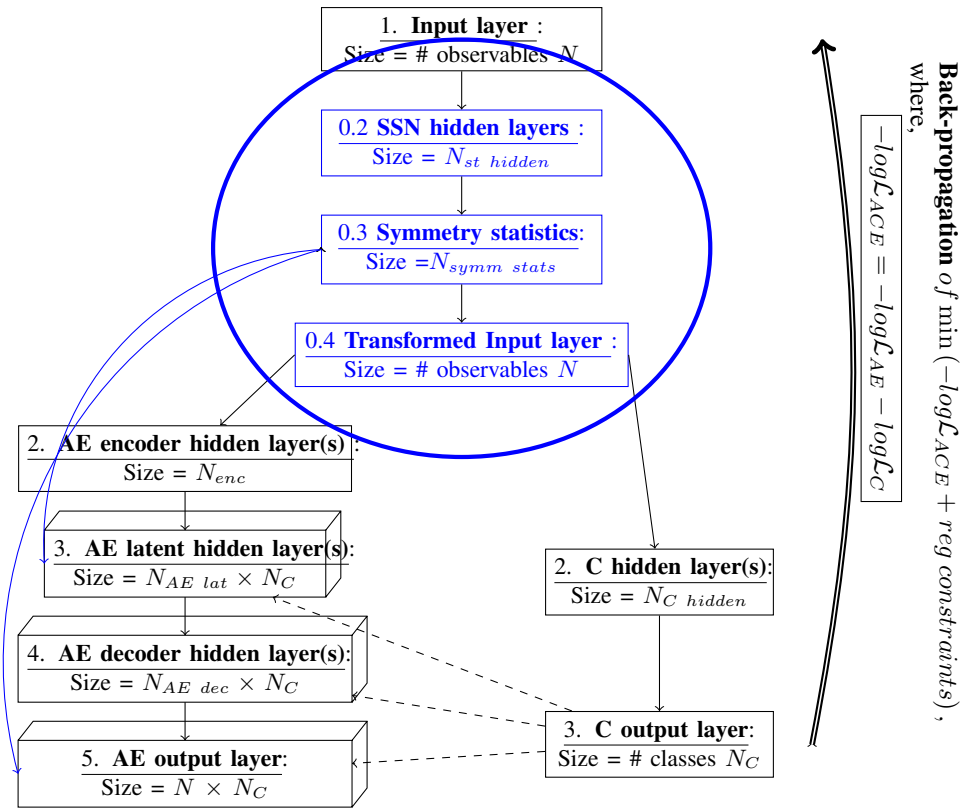
\begin{figure}[!ht]
%\vskip 0.2in
\begin{center}
\begin{tikzpicture}
%\tikzstyle{ann} = [draw=none,fill=none]

\node[rectangle, text width = 0.25\columnwidth, text  centered,  draw ] 
	(Input) {\small \underline{1. \textbf{Input layer} }:\\Size = \# observables  $N$ };
\node[rectangle,  blue, text width = 0.25\columnwidth, text centered, below  =0.2in and  of Input, draw] 	(SSN)  {\small \underline{0.2 \textbf{SSN hidden layers} }: \\Size = $N_{st~hidden}$ };
\node[rectangle,  blue, text width = 0.25\columnwidth,  text  centered, below  = 0.2in  of SSN,draw] 
	(SymmStats)  {\small \underline{0.3 \textbf{Symmetry statistics}}:\\ Size =$N_{symm ~stats}$ } ;
\node[rectangle, blue,text width = 0.3\columnwidth, text  centered, below  =0.2in  of SymmStats, draw ] 
	(TransformedInput) {\small \underline{0.4 \textbf{Transformed Input layer} }:\\Size = \# observables  $N$ };

\node[rectangle,  text width = 0.3\columnwidth, text centered, below  left =0.2 in and -0.3in of TransformedInput,  draw] 
	(AE_EncoderHidden)  {\small \underline{2. \textbf{AE encoder hidden layer(s)} }: \\Size = $N_{enc}$ };
\node[parallelepiped,  text width = 0.25\columnwidth, parallelepiped offset y=0.1in, text  centered, below   = 0.2in  of AE_EncoderHidden,draw] 
	(AE_Latent/Feature)  {\small \underline{3. \textbf{AE latent hidden layer(s)}}: \\Size = $N_{AE~lat} \times N_C$} ;
\node[parallelepiped,  text width = 0.3\columnwidth, parallelepiped offset y=0.1in, text  centered, below  =0.2in   of AE_Latent/Feature, draw] 
	(AE_DecoderHidden)  {\small \underline{4. \textbf{AE decoder hidden layer(s)}}:  \\Size = $N_{AE~dec} \times N_C$} ;
\node[parallelepiped,  text width = 0.3\columnwidth,  parallelepiped offset y=0.1in, text  centered, below=0.2inof AE_DecoderHidden, draw] 
	(AE_Output)  {\small \underline{5. \textbf{AE output layer}}:\\Size = $N \times N_C$};

\path [->](TransformedInput.west) edge node {} (AE_EncoderHidden); 
\path [->](AE_EncoderHidden) edge node {} (AE_Latent/Feature);
\path [->](AE_Latent/Feature) edge node {} (AE_DecoderHidden); 
\path [->](AE_DecoderHidden) edge node {} (AE_Output);

%\node[rectangle,  text width = 0.4\columnwidth,  text  centered, below =0.2in of SymmStats, draw] 
%	(SSTransform)  {\small 4. \textbf{Symmetry statistics}\\ \textbf{Transformation} } ;

\node[rectangle,  text width = 0.2\columnwidth,  text  centered, below right =2.5in and -0.3in of Input, draw] 
	(C_DecoderHidden)  {\small \underline{2. \textbf{C hidden layer(s)}}:\\ Size = $N_{C~hidden}$} ;
\node[rectangle,  text width = 0.2\columnwidth,  text  centered, below=.5inof C_DecoderHidden, draw] 
	(C_Output)  {\small \underline{3. \textbf{C output layer}}: \\ Size = \# classes $N_C$ };

\path [->](Input) edge  node {} (SSN); 
\path [->](SSN) edge  node {} (SymmStats);
\path [->](SymmStats) edge node {} (TransformedInput); 

\path [->](TransformedInput.east) edge  node {} (C_DecoderHidden);  
\path [->](C_DecoderHidden) edge node {} (C_Output); 
	
\path [->](C_Output) edge [dashed] node {} (AE_Latent/Feature); 
\path [->](C_Output) edge [dashed] node {} (AE_DecoderHidden);
\path [->](C_Output) edge [dashed] node {} (AE_Output); 
		
\draw [->]  ([xshift=0.2in]C_Output.east) edge [bend right=10, thick, double]  node [rotate=270, midway, yshift = 0.3in] { $\begin{array}{l} \textbf{Back-propagation } of  \min \left(- log \mathcal{L}_{ACE} + reg~constraints\right),    \\ \text{where, } \\ \qquad  \boxed{- log \mathcal{L}_{ACE} = - log \mathcal{L}_{AE} - log \mathcal{L}_{C} }  \end{array} $} ([xshift=0.in -0.5in + 0.3\columnwidth]Input.east);

%\draw [->]  ([xshift=0.in]SymmStats.east) edge [bend right=90, ultra thick, double]  node [rotate=270, midway, yshift %= 0.in] {  $\begin{array}{c} \small \textbf{Symmetry statistics} \\ \\ \small \textbf{transformation}  \end{array}$} %([xshift=0.in ]TransformedInput.east);

\draw [->]  ([xshift=0.in]SymmStats.west) edge [bend right=45, blue]  node [rotate=270, midway, yshift = 0.in] {} ([xshift=0.in ]AE_Latent/Feature.west);

\draw [->]  ([xshift=0.in]SymmStats.west) edge [bend right=45, blue]  node [rotate=270, midway, yshift = 0.in] {} ([xshift=0.in ]AE_Output.west);

%\node[ellipse, draw=red, ultra thick,  fit=(AE_Latent/Feature) (SymmStats)](FIt1) {};
\node[ellipse, draw=blue, ultra thick,  fit=(TransformedInput)  (SSN) (SymmStats) ](FIt2) {};
        
\end{tikzpicture}
\caption{ACE  architecture \textbf{with symmetry statistics}: compared to the basic generative ACE from \cite{Georgiev15-0}, new components are in blue oval. $\mathbf{AE}$ stands for ``auto-encoder'', $\mathbf{SSN}$ stands for ``symmetry statistics net'', $\mathbf{C}$ stands for ``classifier''.  The arrow from the symmetry statistics to the $\mathbf{AE}$ latent variables indicates that one can sample from the former as well. The arrow from the symmetry statistics to the $\mathbf{AE}$ output layer indicates that one has to invert the transformation from box 0.4, before computing reconstruction error. On the test set, the class probabilities are provided by the classifier as in (\ref{3.1}), hence the dashed lines.}

\label{Fig3.1}
\end{center}
%\vskip -0.2in
\end{figure}

The ACE architecture with symmetry statistics is on Figure \ref{Fig3.1}. As in the basic ACE, training is supervised i.e. labels are used in the auto-encoder and every class has a dedicated decoder, with  unimodal sampling in the generative layer of each class.  The sampling during testing is instead  from a mixture of densities, with mixture weights  $\{\omega_{\mu,c}\}_{c=1}^{N_C}$ for  the $\mu$-th observation, for class $c$, produced by the classifier. The posterior densitiy from section \ref{Generative nets} becomes\footnote{Using a similar mixture of posterior densities, but different architecturally conditional VAEs, were proposed in the context of semi-supervised learning in \cite{Kingma14-3}.}:
\begin{align}
p(\mathbf{z} |\mathbf{x}_{\mu}  )= \sum_{c=1}^{N_C} \omega_{\mu,c} p(\mathbf{z}|\mathbf{x}_{\mu}, c).
\label{3.1}
\end{align}
After interim symmetry statistics are computed in box 0.3 on Figure \ref{Fig3.1}, they are used to transform the input (box 0.4), before it is sent for reconstruction and classification. The inverse transformation is applied right before the calculation of reconstruction error.

Plugging  the symmetry statistics in the latent layers allows to deterministically control the reconstructed observations. Alternatively, sampling randomly from the symmetry statistics, organically ``augments'' the training set. External augmentation is known to improve significantly a net's classification performance \cite{Ciresan12}, \cite{Krizhevsky12}. This in turn improves the quality of the symmetry statistics and creates a virtuous feedback cycle.

%%%%%%%%%%%%%%%%%%%%%%%%%%%%%%%%%%%%%%%%%%%%%%%%%%%%%%%%
\section{Open problems.} 
\label{Open problems} 
\begin{enumerate}[nolistsep]

\item
Test experimentally deep convolutional ACE-s, with (shared) feature maps, both in the classifier and the encoder. From  feature maps at various depths, produce corresponding generative latent variables. Add symmetry statistics to  latent variables at various depths. 
\label{Deconvolution}

\item
Produce separate symmetry statistics for separate feature maps in generative nets, in the spirit of \cite{Hinton11}.
\label{Symm stats per feature}

\end{enumerate}

\subsubsection*{Acknowledgments}
We appreciate discussions with Nikola Toshev, Stefan Petrov and their help with  CIFAR10.

\bibliography{bibliographyNN}
\bibliographystyle{iclr2016_conference}

%Supplementary material
%%%%%%%%%%%%%%%%%%%%%%%%%%%%%%%%%%%%%%%%%%%%%%%%%%%%%%%%%%%%%%%%%%%%%%%%%%%%%%%%%%%%%

\begin{appendices}
\section[]{Implementation.}
\label{Implementation}
All cited nets are implemented on the Theano platform,  \cite{Theano12}. Optimizer is Adam,  \cite{Kingma14-2}, stochastic gradient descent back-propagation, learning rate = 0.0015 for MNIST and 0.0005 for CIFAR10, decay = 50 epochs, batch size = 250. We used only one standard set of hyper-parameters per dataset and have not done hyper-parameter optimizations. Convolutional weights are initialized uniformly in $(-1,1)$ and normalized by square root of the product of dimensions. Non-convolutional weight initialization is as in \cite{Georgiev15-0}.

\textbf{Figure \ref{Fig1.1}}: Auto-encoder branch as in \cite{Georgiev15-0} Figure 9, Gaussian sampling. Classifier branch is convolutional, with 3 hidden layers, with  32-64-128 3x3 filters respectively, with 2x2 max-poolings and a final fully-connected layer of size 700; dropout is 0.2 in input and 0.5 in hidden layers. \textbf{Figure \ref{Fig1.2}}: Same auto-encoder and classifier as in Figure \ref{Fig1.1}. A symmetry statistics \emph{localization} net, as in \cite{Jadeberg15}, produces six affine spatial symmetry statistics (box 0.2 in Figure \ref{Fig3.1}). This net has 2 convolutional hidden layers, with  20 5x5 filters each, with 2x2 max-poolings between layers, and a fully-connected layer of size  50. \textbf{Figure \ref{Fig1.3}}: Layer sizes 3072-2048-2048-(2x10)-(2048x10)-(2048x10)-(3072x10) for the auto-encoder branch, same classifier as in Fig \ref{Fig1.2}. The symmetry statistics net has 2 convolutional hidden layers, with  32-64 3x3 filters respectively, with 2x2 max-poolings between layers, and a fully-connected layer of size 128. \textbf{Figure \ref{Fig1.4}}: Two convolutional layers  replace the first two hidden layers in the encoder, with  32-64 5x5 filters respectively. The two corresponding deconvolution layers are at the end of the decoder. Layer size 2048 is reduced to 1500 in the auto-encoder, Laplacian sampling, rest is the same as in Figure \ref{Fig1.3}.

\section[]{Distorted MNIST.}
\label{Distorted MNIST}
The two distorted MNIST datasets replicate \cite{Jadeberg15}, Appendix A.3, although different random seeds and implementation details may cause differences. The rotated-translated-scaled (RTS) MNIST is on 42x42 canvas with random +/- 45$^{\circ}$ rotations, +/- 7 pixels translations and $1.2$/$0.7$ scaling. The translated-cluttered (TC) MNIST has the original image randomly translated across a 60x60 canvas, with 6 clutter pieces of size 6x6, extracted randomly from randomly picked other images and added randomly to the background.
%\begin{figure}[!ht]
%%\vskip 0.2in
%\includegraphics[width=\columnwidth]{Test200_MNISTandDeformedMNIST.png}
%\caption{Two hundred observations from the respective test sets of each of the three MNIST data sets under consideration. \textbf{Top Left.} Original MNIST, on 28x28 canvas. \textbf{Top Right.} Rotated-translated-scaled (RTS) MNIST, on 42x42 canvas. \textbf{Bottom.} Translated-cluttered MNIST, on 60x60 canvas.}
%\label{FigA.1}
%%\vskip -0.2in
%\end{figure} 

\end{appendices}
\end{document}